\documentclass[sigconf]{acmart}

\AtBeginDocument{%
  \providecommand\BibTeX{{%
    \normalfont B\kern-0.5em{\scshape i\kern-0.25em b}\kern-0.8em\TeX}}}

\setcopyright{acmcopyright}
\copyrightyear{2023}
\acmYear{2023}

\acmConference[MLoG@WSDM'23]{ACM Conference}{March 2023}{Singapore}%

\acmPrice{15.00}
\acmISBN{978-1-4503-XXXX-X/18/06}

\usepackage{graphicx} 

\usepackage{algorithm}
\usepackage{color}

\usepackage{titlesec}

\usepackage{tabulary}
\usepackage{amsmath}
\usepackage{graphicx}
\usepackage{multirow,comment,bbm}
\usepackage{amsthm}
\usepackage{xcolor}

\usepackage{mathtools}

\newtheorem{theorem}{Theorem}

\newtheorem{lemma}{Lemma}
\theoremstyle{definition}
\newtheorem{definition}{Definition} 
\theoremstyle{remark}
\newtheorem{remark}{Remark}
\theoremstyle{definition}
\newtheorem{problem}{Problem}

\newcommand\harp[1]{\mathstrut\mkern2.5mu#1\mkern-11mu\raise0.6ex%
	\hbox{$\scriptscriptstyle\rightharpoonup$}}

\newcommand{\mc}[1]{\mathcal #1}
\newcommand{\mb}[1]{\mathbf #1}
\newcommand{\mbb}[1]{\mathbb #1}

\newcommand{\subidxU}{{u\!}}

\newcommand{\paper}{paper}

\newcommand{\edgeprobmatrix}{\mb P}
\newcommand{\dirimages}{./}

\newcommand\Ttwod{\mc T}
\newcommand\rv{random variable}

\newcommand\inputgraph{\mc G}

\newcommand\modelfunction{f_{G}(\Theta)}
\newcommand\modelfunctionNoTheta{f_{G}}

\newcommand\approxjointmodelfunction{\tilde{\mc F}_{\mc G}(\Theta)}
\newcommand\jointmodelfunction{\mc F_{\mc G}(\Theta)}

\newcommand\modelfunctionParam[1]{{f_{G}}_{#1}(\Theta)}

\newcommand\mainstatistic{\phi}
\newcommand{\exchangeability}[0]{exchangeability}
\newcommand{\exchangeable}[0]{edge-exchangeable}
\newcommand{\UpExchangeability}[0]{Exchangeability}

\usepackage{nicefrac}

\newcommand{\secLabelBackground}{Preliminaries}
\newcommand{\scalefree}{power-law}

\begin{document}

\title{Goodness-of-Fit 
of 
Attributed Probabilistic\\Graph Generative 
Models}


\author{Pablo Robles-Granda}
\email{pdr@illinois.edu}
\affiliation{%
  \institution{University of Illinois at Urbana-Champaign}
  \country{USA}
}

\author{Katherine Tsai}
\email{kt14@illinois.edu}
\affiliation{%
  \institution{University of Illinois at Urbana-Champaign}
 \country{USA}
}

\author{Oluwasanmi Koyejo}
\email{sanmi@stanford.edu}
\affiliation{%
  \institution{Stanford University}
 \country{USA}
}

\renewcommand{\shortauthors}{Robles-Granda et al.}


\begin{abstract} 

Probabilistic generative models of graphs are important tools that enable representation and sampling. 
Many recent works have created probabilistic models of graphs that are capable of representing not only entity interactions but also their attributes. 
However, given a generative model of random attributed graph(s),  the general conditions that establish goodness of fit are not clear a-priori. In this paper, we define goodness of fit in terms of the mean square contingency coefficient for random binary networks. For this statistic, we outline a procedure for assessing the quality of the structure of a learned attributed-graph by ensuring that the discrepancy of the mean square contingency coefficient (constant, or random) is minimal with high probability. We apply these criteria to verify the representation capability of a probabilistic generative model for various popular types of graph models. 

\end{abstract}





\maketitle

\section{Introduction}
\vspace{-1mm}

\emph{Labeled graphs} are powerful tools to represent complex systems components and their interactions 
\cite{Myunghwan:10,Spyropoulou:14}. 
For instance, 
metabolite types
in metabolic networks, political affiliation in social networks, and behavior types in a network of birds, can all be modeled as node-attributes  \citep{Zhou:11,Ahmed:09,Jeong:00,Psorakis:12}. Further, the properties of many real networks include community structure with connections drawn from a \emph{\scalefree~degree} distribution.
Models such as the preferential-attachment model \citep{BarAlb99}, the cumulative-advantage model \citep{de1976general}, the Holme-Kim model~\citep{holme2002growing}, among others, 
generate graphs with power-law degree distributions.

While node attributes contain insightful information on the properties of elements linked by the underlying graph structures, modeling the associations of node attributes and graph structures is a challenging problem.
To address this issue, 
one can construct hierarchical Probabilistic Generative Models (PGM) by modeling 
the marginal distributions of the node attributes and edges' structure~\cite{Pfeiffer:14}. This approach simplifies the procedure of data fitting and avoids cyclic dependencies. 
Our work particularly focuses on the generative modeling of binary attributes.
Since modeling attributes is complex (due to the attribute types: discrete vs. continuous attributes; the mathematical representation; and data size and dimensionality), our work creates a 
framework but focuses on binary attributes.
Despite the numerous contributions to attributed-graph modeling
\cite{bothorel:2015,Silva:2012,Kolar:2014}, 
it remains unclear what are the general conditions that guarantee a generative model from attributes can capture the true generative process of graph(s) nor is it clear how to assess 
the goodness-of-fit from underlying graph distributions.

While goodness-of-fit measures for graphs is a thriving area of
research \cite{pmlr-v80-yang18c,leppala2017admixturegraph, chen2019bootstrap,SHORE201516,weckbecker2022rkhs},  
goodness-of-fit for attributed
graphs is less explored \cite{faloutsos2019social,adriaens2020block}. 
Existing work \cite{dimitriadis2020triage} relies on
traditional metrics such as the R-Squared. We
define characteristics of the parameters that specify when
structure and node-attributes are captured simultaneously, as
opposed to separately through traditional metrics.

We focus on probabilistic generative models of binary 
attributed graphs. Under this setting, we identify that the mean square contingency coefficient~\citep{cramer1946mathematical} can be used to assess the quality of the representation of attributed graphs.
We developed a theoretical framework to understand generative models of complex graphs guided by a statistic of the data and the model. Specifically, we choose models that minimize the distance of these statistics as measured by the mean contingency coefficient from the data vs. the one that may be derived from graphs from the model. 
%
Our 
contributions are the following: 
we (1) formalize the goodness of fit measure for labeled graphs and establish its characteristics in the parameter space; (2) derive the mathematical conditions necessary to ensure the faithful representation of the graph data with high probability 
(3) evaluate this framework empirically on various existing and widely used generative models of graphs where labels are incorporated. 
\vspace{-5mm}
\subsection{Problem Description}

Let $G=(\mathbf V,\mathbf E)$ be a graph with set of vertices $\mathbf V$ and edges $\mathbf E \subset \mathbf V \times \mathbf V$. We define $\mc A_{ij}$ to be a binary random variable, where its realization $A_{ij}\!=\!1$ indicates that the edge $e_{ij}$ between nodes $V_i,\!V_j \in \mathbf V$ exists ($e_{ij}\in\mathbf E$), and $A_{ij}\!=\!0$ if $e_{ij}\!\notin\!\mathbf E$. 
Thus, $A$ is an adjacency matrix. 
We denote $\modelfunction: \mathbb R^{d}\rightarrow [0,1]^{|V|\times|V|}$ (where $d \in \mathbb{N}$) as a probabilistic generative model of graphs (PGM) with parameter $\Theta$ that generates a network $G$ through a sampling process.
The process is represented by using a $|\mathbf V|\!\times\!|\mathbf V|$ probability matrix $\edgeprobmatrix$, where $\edgeprobmatrix_{ij}\!=\!Pr(\mc A_{ij}=1)$ is the probability of an edge between $V_i$ and $V_j$. The random variables $A_{ij}$ may not be independent. 

Given the \rv~$D_i=\sum_{j=1}^{|\mathbf V|} \mc A_{ij}$, a PGM of scale-free graphs samples graphs with 
degree density function
$f_{D_i}(d)\sim d^\gamma$ for some $\gamma>0$. 
In the following, definitions \emph{with high probability} means 
\emph{with
probability greater than $1-\delta$} for some small value $\delta>0$.

 We define ${\bf X}=\{X_{1},$ $\ldots,X_{|\mb V|}\}$ be the node attributes for a graph $G$. Let $\{\mathcal\inputgraph_i\}_{i=1}^{N_g}$, where $\mathcal\inputgraph_i=(G_i,\mathbf X_i)$ for $i=1,\ldots, N_g$, be a set of input attributed graph(s) and $\jointmodelfunction$ be the joint distribution of both the graph and the node attributes. 
We denote $\mc S$ as the statistic that measures the label-structure dependencies of $\edgeprobmatrix$ and $\mb X$.

 In our work, we are interested in the capability of a model to achieve representation of a graph, namely representing not only their attribute and graph distribution but their interaction $\mc S$ (see formal Def. \ref{definition:graph_representation}). We now formalize the problem of interest. 
\begin{definition}[Representation]
Given an 
attributed graph $\mathcal\inputgraph_0$, 
we say that $\mathcal\inputgraph_0$ is representable by a probabilistic model 
$\jointmodelfunction$ with respect to a graph 
statistic $\mathcal S$ 
if
the absolute difference between the sample statistic 
$\mc S(\mc\inputgraph_0)$ and the statistic $\mc S(\mc G_i)$ from any random graph $\mc G_i$ sampled from 
$\jointmodelfunction$ 
converges to $0$ with high probability. 
\end{definition}


\begin{problem}[Conditions for Representation] 
Given an attributed graph, $\mathcal\inputgraph_0$ an a candidate model $\jointmodelfunction$ our objective is 
to identify the properties of $\jointmodelfunction$ s.t., $\mc G_0$ is representable by $\jointmodelfunction$ with respect the some graph statistic $\mc S$. %
\end{problem}

For the choice of $\mc S$ we must use  
a function (or statistic) that captures interactions of graph structure and node attributes. 
%
In consequence, the \emph{properties} of $\jointmodelfunction$ are nothing but 
the parameters 
and structural requirements to guarantee sampling graphs given a choice of statistics, i.e., $\mc S$. Thus, our Problem characterizes the conditions for the convergence defined as \emph{representation} of a graph.

In practice, we use the mean square contingency coefficients (MSCC), denoted as $\mainstatistic$, as the statistic $\mc S$ to evaluate the difference between $\mc S(\mc G_0)$ and $\mc S(\mc G_i)$. 
We chose MSCC because it has several advantages over comparable measures. It is more robust to inbalanced scenarios \cite{powers2020evaluation} than others, with some limitations \cite{zhu2020performance}, but at the same time behaves similarly to Pearson correlation in the case of binary variables \cite{cramer1946mathematical,powers2020evaluation}. 
Then, to solve our problem we derive the probability of sampling graphs with a chosen $\mainstatistic$. 
Our contributions are as follows: 
(a) we prove that for a generative model
of graphs $\modelfunctionNoTheta$ 
and the binary 
multivariate attributes ${\mb{X}}$, 
one can verify the size of graphs for which $\modelfunctionNoTheta$ can learn the attribute-structure interactions with high probability. %
(b) we identify the parametric conditions of $\modelfunctionNoTheta$ (in terms of the statistic $\mc S$) 
that guarantee the target $\mainstatistic$ can be obtained independently of the value of the target $\mainstatistic$; 
(c) We use our formulation as a goodness-of-fit measure to perform model selection on Stochastic Block Models \cite{holland1983stochastic} in the experimental evaluation. %
(d) Finally, we use our formulation for modeling the attribute-feasibility of several generative models of graphs.

\vspace{-1mm}
\section{Preliminaries}
\label{sec:bgn}
\vspace{-1mm}

\noindent\textbf{
Probabilistic
Models of Scale-Free Attributed-Graphs.}

We consider the following generative model, denoted as $\approxjointmodelfunction$, to approximate true $\jointmodelfunction$. First, we sample the vector of attributes for some fitted attribute distribution 
$f_X(x)$ and then sample candidate $\mc G 
$ from 
a
proposal conditional distribution $f_{G|X}(\Theta)$:
\begin{align}\label{eq:approx_pgm}
    \mb X&\sim f_X(x);\notag\\
    G|\mb X &\sim f_{G|X}(\Theta),
\end{align}
Third, we accept the candidate $\mc G$ with some probability $p_{{\mc C}}$ determined by the proposal distribution. 
We assume marginal distributions 
the same as $\modelfunctionParam{}: \mathbb{R}^d\rightarrow [0,1]^{|V|\times|V|}$ (for $d \in \mathrm{Z}^+$). 

This approximation schema is general and able to incorporate characteristics that most real-world graphs have, including scale-free degrees 
\cite{crane2016edge},
sparsity,  \exchangeability, attribute-correlation preservation \citep{Bansal:2004}, projectivity \citep{shalizi:2013}, and others.
Intuitively, a sequence of random graphs is \exchangeable~when the underlying generative distribution is invariant to finite permutations of the edge realization process. We now introduce the formal definition adapted from \citet{cai:2016}. Consider the superindex $i$ as an indicator of the graph associated to the edges $\mb E$, and  $k\in [n_s], n_s< {|\mb V|}^2$ is the iteration of the edge-exchangeable sampling process.

\begin{definition}[\UpExchangeability]
Let $\sigma$ be a permutation of integers in $[n_s]$.  For an 
edge set in each $G_i$, a random graph generator of the sequence $(G_i)_{i\in \mathrm{N}}$ 
is parameter-wise infinitely \exchangeable~if for every $i,j\in \mathrm{N}$ and every $\sigma$, then $G_i\overset{d}{=} G_j$ for $|\mb V_i|=|\mb V_j|$, i.e., the joint distributions of candidate edges are equal $Pr(\mb E^i_1,\ldots,\mb E^i_{n_s})=Pr(\mb E^j_1,\ldots,\mb E^j_{n_s})$ and $\mb E^i_{\sigma(k)}=\mb E^j_k$.
\end{definition}
\vspace{-1mm}
In our work, we are interested in the adjacency matrix directly to avoid sampling graphs 
and to simplify the process to combine attributes and structure. Thus, effectively evaluating the model without the sampling process.
We will see later that an application of our framework is the  computation of test statistics for which generation of graphs is not needed. Hence, we make the following assumptions.

\noindent\textbf{Assumption 1}\emph{ (Sampling-agnostic \exchangeability~of structure)} Any graph model $\modelfunction$, independently of the 
geometric realization of the graphs, 
must be \exchangeable~and, thus, it must guarantee sparsity and sampling stationarity \citep{cai:2016}.

\noindent\textbf{Assumption 2}\emph{ (\UpExchangeability~preservation)}
An approximation $\approxjointmodelfunction$ should maintain the edge exchangeability defined by $\modelfunction$.

There are various graph models that satisfy Assumption 1 -- 2, including  the Erd\H{o}s-R\'enyi model (ER) \cite{renyi1959random}, the Stochastic Block Model (SBM) \cite{holland1983stochastic}, and the Graph Frequency Model (GF) \cite{cai:2016}.

Some models of scale-free graphs are \emph{mechanistic}, 
since
they rely on an iterative algorithm to add edges to the sampled graph in a way that ensures the power-law of degree distributions, and other characteristics \cite{BarAlb99,de1976general,holme2002growing}. Thus, do not satisfy the \exchangeable\ property stated in Assumption 1 -- 2. %

In the following, we discuss examples of ways that attributes could interact with the graph, under the model~\eqref{eq:approx_pgm}. This builds the foundation of our analysis in the following section. Consider the PGM $\modelfunctionNoTheta$ and the binary 
set of (data) attributes ${\mb{X}}=\{\mb X^1,\ldots,\mb X^m\}$ and (output) sample attributes $\tilde{\mb{X}}=\{\tilde{\mb X}^1,\ldots,\tilde{\mb X}^m\}$ from a fitted probability mass function. 
%
We make the following observations:

\noindent\textbf{Observation 1} \emph{ (Effect of $\tilde{\mb X}$ on $\tilde G$ - sample)} The elements of an attributed graph interact in two ways: (1) $\tilde{\mb X}$ labels every entry of $\tilde{\mb V}$, thus for every pair of nodes $(\tilde{\mb v}_i,\tilde{\mb v}_j)$ s.t. $\tilde{\mb v}_i, \tilde{\mb v}_j\in \tilde{\mb V}$,  the pairs $(\tilde{\mb x}_i^p,\tilde{\mb x}_j^q)$ for $p,q \in [1,m]$ form potential edge labels of $\tilde G$. (2) $\tilde{\mb X}$ labels every entry of $\tilde{\mb E}$, thus, for every pair of nodes $(\tilde{\mb v}_i,\tilde{\mb v}_j)$ s.t. $\tilde{\mb e}_{i,j}\in \tilde{\mb E}$  there are pairs $(\tilde{\mb x}_i^p,\tilde{\mb x}_j^q)$ for $p,q \in [1,m]$ that are actual edge labels of $\tilde G$. 

There might be repeated edge-labels in both cases. Hence, we will represent the unique list of edge-labels as $\Psi$ (for example $\Psi=\{00,10,11\}$ for undirected graphs with binary attributes and assume the value is homogeneous between data and sample.

\noindent\textbf{Observation 2} \emph{ (Effect of $\mb X$ on $\mb P$ - data/model)} 
%
The pair $(\edgeprobmatrix, \mb X)$ can be 
also 
represented with the pair $(\mathbf{U},\Ttwod)$, where
$\mathbf U=\{\pi_1,\pi_2,\ldots,\pi_\subidxU~,\ldots,\pi_{\kappa}\}$ is the set of unique Bernoulli parameters that appears in the matrix $\edgeprobmatrix$ ($\kappa=|\mathbf U|$), i.e. $\mathbf U={\Phi}_{\Theta}(\edgeprobmatrix)$. The function ${\Phi_{\Theta}}$ factorizes $\edgeprobmatrix_{ij}$ into its parametric components, and hence depends on $\modelfunctionNoTheta$. 
$\Ttwod$ is a matrix where each entry contains a set of positions $\Ttwod_{j,\subidxU}$ for pair of nodes with labels $\Psi_j$ and probability $\pi_\subidxU$ of link between them.

\noindent\textbf{Observation 3} \emph{ (Attribute-structure interactions)}
 Consider the input $\mb X,G$ and construct the vectors $\mc X^p,\mc X^q$ s.t. $\mc X^p=\{\mb x^p_i\}$ and $\mc X^q=\{\mb x^p_j\}$ for each $\mb e_{i,j}\in \mb E$. 
From Observation 1--2, we can infer that the 
interactions of 
$\mb X,G$ 
can be summarized with 
${\beta},|\mb E|$
, where the j-th entry of $\beta$, denoted as 
$\beta_j$,  is the fraction of edges that share the same label $\Psi_j$. %
A similar observation applies to the sample $\tilde{\mb X},\tilde{G}$.

%

\section{Main Results -- Representation Closeness}
\label{Sec:Theory}
\vspace{-1.5mm}

We begin this section by first introducing the notion of MSCC and notations. Then, we introduce Definition~\ref{definition:graph_representation} to mathematically formulate Problem 1 using the MSCC. 
Then, to solve Problem 2, we use Theorem~\ref{thm:mainthm},  \ref{thm:mainthmrnd} which provides a relation between the size of the candidate edges of $\modelfunction$ as a consequence of $\jointmodelfunction$.

 Our work is a generalization of the work of \citet{davenport51sanhury} in the case of the $\phi$-coefficient associated with random graphs:
 \begin{definition}[{Mean Square Contingency Coefficient}] The $\phi$-coefficient is a measure of association between two binary variables. In a contingency table with entries $n_{ij}$ for $i,j\in \{0,1\}$,  $\phi$ is defined as $\phi = \frac{n_{11}n_{00}-n_{10}n_{01}}{\sqrt{n_{1\bullet}n_{0\bullet}n_{\bullet0}n_{\bullet1}}}$. 
 \end{definition}
 Notice that the coefficient can be defined for any categorical variables and the size will be $p m$ for variables with cardinality of categories $p$ and $m$, respectively. The bullet $\bullet$ indicates all the rows/columns (i.e., the total for the column/row, respectively). 
 
 In the case of contingency tables of size $2\times 2$, $\phi$ is equivalent to the Pearson correlation $\rho$, which is why we use this to simplify our analysis. 
Our work is a generalization of (\citet{davenport51sanhury}; 1991) for the case of tables derived from random graphs.
We compute the $\phi$ coefficient globally and reformulate the $n_{ij}$ entries as $\beta_{k}$ as detailed in the following notation description. These values will be computed for both the original graph data and for graphs sampled from the models. 

\noindent\textbf{Notation For Theorems and Lemmas.}
We summarize and introduce additional notation for the lemmas and theorems. 
We denote as $\edgeprobmatrix$ the matrix of edge-probabilities of a structural model and $\mb X$ the attribute-values associated with the nodes. $\Ttwod_{ij}(\Psi_i,\pi_j)$ is the list of possible edges associated with parameter $\pi_j$ and edge-type $\Psi_i$ (described in \secLabelBackground).
We denote as $\mathbf U=\{\pi_j\}_{j=1}^{\kappa}$, the set of unique probabilities from probabilistic generative model $\mc M$.

Let the edge-types $\Psi=\{00,01,11\}$ come from Bernoulli-distributed node-attributes $\mathbf X$ in an undirected network. Let $N_{i,j}=|\Ttwod_{ij}|$ be the cardinality of the list of possible edges associated with parameter $\pi_j$ and edge-type $\Psi_i$, 
$n_j=\sum_{i=1}^{|\Psi|} N_{ij}$ be the total number of possible edges per $\pi_j$, $r_j=(N_{1j}+N_{3j})/(\sum_{i=1}^{|\Psi|} N_{ij})$ be the fraction of possible edges of type $\{11,00\}$, 
and $y_j\sim Bin(n_j,\pi_j)$ be the number of edges to sample. Under the condition of $\Psi$, the MSCC $\phi$ is equivalent to the correlation $\rho$. Hence, for the main theorems we will focus on the values target correlation we desire to model (data) and the output correlation of the sampled graph, denoted as $\rho_{IN},\rho_{OUT}$ respectively. For binary labels $\rho_{IN},\rho_{OUT}$ can be represented as $\beta,\tilde \beta$ in terms of edge-labels as described in the theorems, explained in Observation 3.

We now formally define the notion of graph representation: 

\begin{definition}\label{definition:graph_representation}{ (Graph Representation - Formal Definition)}
Given the input graph $\mc G=\{G, \mb X\}$ and an output graph $  \tilde{\mc G} =\{ \tilde{G}, \tilde{\mb X}\}$ sampled from 
$\approxjointmodelfunction$ fitted by the data. Then, %
 we say that $\approxjointmodelfunction$ is an $\epsilon$-representation of $\mc G$ 
if  
 $|\rho_{_{IN}}-\rho_{_{OUT}}|<\epsilon$ for small $\epsilon>0$. 
\end{definition}%

Lemma \ref{thm:corr-bound} provides a criterion for boundedness of the correlation in terms of 
$r_j$
per parameter $\pi_j$. It tells us the condition when there is an upper-bound of $\rho_{OUT}$ beyond which input data cannot be represented:

\begin{lemma}
\label{thm:corr-bound}{(Boundedness of Representation)} %
Let $D_{KL}(r_j||\pi_j)$ be the Kullback-Leibler divergence of Bernoulli distributions with parameters $r_j$ and $\pi_j$, and $c_0=23.03$ be a universal constant.
If there exists $\pi_j\in \mathbf U$ such that $n_jD_{KL}(r_j||\pi_j)\geq c_0$, for $c_1=1-10^{-10}$, then
\noindent a) if $0< r_j <\pi_j$, then
$\rho_{OUT}\leq c_1$ and $c_1<1$;
\\
b) if $\pi_j< r_j<1$, then
$\rho_{OUT}\leq c_1$ and $c_1=1$.
\\
\noindent 
\end{lemma}
\vspace{-5mm}
The proof is available in the appendix.
\begin{remark}
Recall from Observation 2 that the sampling of edges 
is done on the binomials defined by $\mb U$, $N,\psi,\beta$ and the edges are indexed by $\mc T$. Thus, the lemma above has important implications.  
Condition (a) implies that %
we must sample %
edges linking nodes with opposite labels $(01)$ because the number of edges needed to sample $y_j$ is \emph{likely} to be greater than node pairs with positively correlated labels. Therefore, there is an upper-bound of the correlation that can be achieved. 
The magnitude of $y_j - N_{1j} - N_{3j}$ determines the maximum achievable correlation, i.e., $\rho_{OUT}<1$. 
Condition (b) implies that it is possible to sample edges to obtain the correlation among them ($\rho_{OUT}$ can be up to $1$), because the number of edges to sample $y_j$ is less than the positively correlated available.
\end{remark}
\begin{remark}
This result applies to any sampling method that draws edges randomly from $P(G)$. 
\end{remark}
The following theorem tells us the probability that the input data can be represented and sampled from a learned model, i.e., $Pr(|\rho_{IN}-\rho_{OUT}|<\epsilon)$.

\begin{theorem}[Correlation Recovery - Constant  $\rho_{IN}$]
\label{thm:mainthm}
Let $\chi_{ij}$ be the number of edges sampled by $\mc S$ per edge-label $\psi_i$ and parameter $\pi_j$
and $\mu=\sum_j^{\kappa} y_j$.
Then, for
any $\mc M$ and $\mc S$ and small $\delta,\epsilon_1,\epsilon_3>0$, the bound of the difference between the target correlation $\rho_{IN}$ and the correlation of the sampled graph $\rho_{OUT}$
has probability 
\begin{equation}
Pr(|\rho_{IN}-\rho_{OUT}|<\epsilon)>
1-\delta,
\end{equation}

for 
$\delta=\sum_{i=1}^{2} \tau_i + \prod_{i=1}^{2} \tau_i$,\\
where
$\tau_i = \exp{\left(-{{\left({(\beta_{i}-\epsilon_i)^{-1}{\sum_{j}^{\kappa} \chi_{ij}}-\mu}\right)}^2}/{3\mu}\right)}+$\\ \hspace*{\fill}
$\exp{\left(-{{\left({\mu}-(\beta_{i}+\epsilon_i)^{-1}{\sum_{j}^{\kappa} \chi_{ij}}\right)}^2}/{2\mu}\right)}$\\
and $\epsilon=
\frac{\beta_3-(p+\Delta p)^2}{(p+\Delta p)(1-p-\Delta p)}
-
\frac{\beta_3-p^2}{p(1-p)}$, 
where $p=\beta_3+\frac{1-\beta_1-\beta_3}{2}$ and $\Delta p=\epsilon_3+\frac{-\epsilon_1-\epsilon_3}{2}$.
\end{theorem}

\begin{remark}
This theorem determines the probability that certain configuration of structure and labels of a reference graph could be sampled for a given estimated model.
Thus, the theorem can be used to verify whether sampling certain number of edges 
will lead to a correlation that is close to the target with high probability. 
This could be useful for post estimation tasks, such as model selection, goodness-of-fit test, sensitivity analysis, 
and others, where assessment of the model is necessary.
This theorem %
shows that to 
compute this probability we can determine $\epsilon$ as the maximum difference in MSCC for small changes in $\beta$, namely  $\epsilon_1,\epsilon_3>0$, which is otherwise not feasible due the degrees of freedom of $\phi$.
\end{remark}

\subsection{Proof of Theorem \ref{thm:mainthm}}

To prove this theorem, first we need some intermediate lemmas to prove differences in MSCC 
can be expressed in closed form. Consider the partial order in $\mbb R^2$ defined as $\Upsilon=\{x\preceq y$ iff $x_i\leq y_i$ for $i=\{1,2\}\}$ and $\Upsilon\subset \mbb R^2
\times \mbb R^2$
and let 
$\mc U=\{(x,y)\in [0,1]\times[0,1]:x+y\leq 1\}$.

\begin{remark}{(Correlation Identities)}
\label{rem:corr_identities}
The 
correlation can equally be expressed as either:
\begin{align}
   &\rho(\beta_1,\beta_3)=\frac{2\beta_1\beta_3+2\beta_1+2\beta_3-\beta_1^2-\beta_3^2-1}{(1-\beta_1+\beta_3)(1+\beta_1-\beta_3)};\label{eq:rho_1}\\
   &\rho(\beta_1,\beta_3)=\frac{\beta_3-p^2}{p(1-p)}\label{eq:rho_2},
\end{align}
where $p=\beta_3+\frac{1-\beta_1-\beta_3}{2} $.
\end{remark}

To prove these identities, we can replace the values of $\vec\beta=[\beta_i]_{i=1}^{|\Psi|}$ in the definition of the 
correlation:
\begin{align*}
    \rho&=%
    \frac{|E|^2\left(\beta_3-\left(\beta_3+\frac{1-\beta_1-\beta_3}{2}\right)^2\right)}{|E|^2\left(\beta_3+\frac{1-\beta_1-\beta_3}{2}\right)\left(1-\left(\beta_3+\frac{1-\beta_1-\beta_3}{2}\right)\right)}=\eqref{eq:rho_2}\\
    &=\frac{(4\beta_3-(1+\beta_1^2+\beta_3^2-2\beta_1+2\beta_3-2\beta_1\beta_3))/4}{(1-\beta_1+\beta_3)/2 (1+\beta_1+\beta_3)/2}=\eqref{eq:rho_1}.
\end{align*}

\begin{definition}
\label{rem:realbimonotonic}
Any function $\gamma\!:\!\mbb R^2\rightarrow \mbb R$ is monotonic with respect to $(\mbb R^2,\Upsilon)$ if %
$x\preceq y $ implies that $\gamma(x)\leq \gamma(y)$ for any $x,y\in\mathbb{R}^2$.
\end{definition}

In other words $\gamma$ is monotonic with respect to the projections along each dimension of its domain.

\begin{lemma}%
\label{thm:monincreasing}
The %
correlation $\rho$ of any binary variable is monotonically increasing with respect to the poset $\Delta_\epsilon$ defined as the pair $(\mc U,\Upsilon)$.

\end{lemma}

The proof of this lemma is provided in the Appendix.  
A sketch proof consists in the following: proof there are dimensions along which there is monotonic increase; then, use Remark \ref{rem:realbimonotonic} to prove monotonic increase in $(\mc U,\Upsilon)$.

\begin{lemma}
\label{thm:maxepsilon} Let $\rho_{OUT}=\rho(\tilde\beta_1,\tilde\beta_3)$ and $\rho_{IN}=\rho(\beta_1,\beta_3)$ be the correlation of the sampled graphs and the correlation of the input graph, respectively. %
Given small $\epsilon_1>0,\epsilon_3>0$, 
the maximum difference $\epsilon=\max(|\rho_{IN}-\rho_{OUT}|)$ that satisfies $|\beta_{1}-\tilde\beta_{1}|<\epsilon_1 \mbox{  and  } |\beta_{3}-\tilde\beta_{3}|<\epsilon_3$
is given by

$\epsilon=
\frac{\beta_3-(p+\Delta p)^2}{(p+\Delta p)(1-p-\Delta p)}
-
\frac{\beta_3-p^2}{p(1-p)}$, 
where $p=\beta_3+\frac{1-\beta_1-\beta_3}{2}$ and $\Delta p=\epsilon_3+\frac{-\epsilon_1-\epsilon_3}{2}$.
\end{lemma}

The proof of this lemma is provided in the Appendix. 
A sketch proof consists in the following: Reformulate the values of $\tilde\beta_i$s in terms of $\beta_j$s and $\epsilon_i$; find an expression for $\max(|\rho_{IN}-\rho_{OUT}|)$ using Lemma \ref{thm:monincreasing}.

Equipped with Lemmas \ref{thm:maxepsilon} and \ref{thm:monincreasing}, we can prove Theorem \ref{thm:mainthm}. The detail is available in the Appendix. 
As a sketch of the proof, the steps include: Consider defining the correlation in terms of $\beta$; identify bound types; use Lemma \ref{thm:maxepsilon} to find $\epsilon$. Identifying the probability of closeness of $\rho$ (data vs. model) using the neighborhood $\epsilon$. $\square$

Our analyses state, for given values of the parameters of the model $\mathbf{P}, \mathbf{X}$, whether the probability that the correlation of a graph %
could be sampled is large enough, i.e., they state if the graph(s) with specific correlation can be sampled, or equivalently when the approximation $\approxjointmodelfunction$ is close to the true $\jointmodelfunction$.

These apply to any model 
and relate $\mathbf P, \mathbf X$ to the probability of modeling/sampling certain type of networks. In summary, our approach can be applied to determine the probability that certain graph structural properties (given degree, connected components, cycles, autocorrelation, etc.) can be sampled for a given estimated model, which simplifies post estimation tasks, including goodness-of-fit test (e.g. Kolmogorov-Smirnov, Anderson-Darling, AIC, etc.). We illustrate this with an application example in 
Section  \ref{sec:prediction-example}.

\section{Empirical Evaluation}

We evaluated our theoretical insights to identify the structural constraints on real world data using the Stochastic Block Model (SBM) and measured feasibility of representation on four graph models: the Erd\H{o}s-R\'enyi model (ER), the Stochastic Block Model (SBM), the Stochastic Kronecker Graph with mixing, and the Graph Frequency Model (GF).

\subsection{Model Selection in Real World Networks}
\label{sec:prediction-example}
\begin{figure}[t]
\begin{center}
\includegraphics[width=9cm]{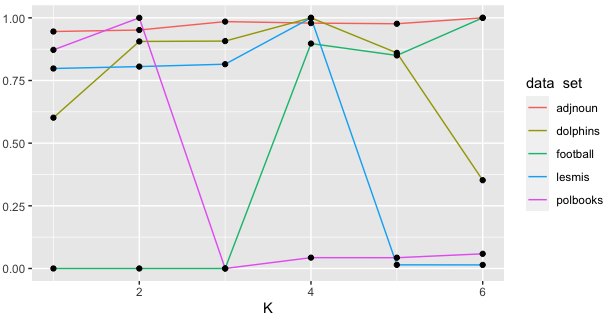}
\vspace{-8mm}
\caption{{\small Model Selection: $K$ vs. probability of representation. Optimal $K$ corresponds to $K$ with maximal score.}}
\vspace{-1mm}
\label{fig:modeling2}
\end{center}
\vspace{-5mm}
\end{figure}

\begin{figure}
\begin{center}
\includegraphics[width=8cm]{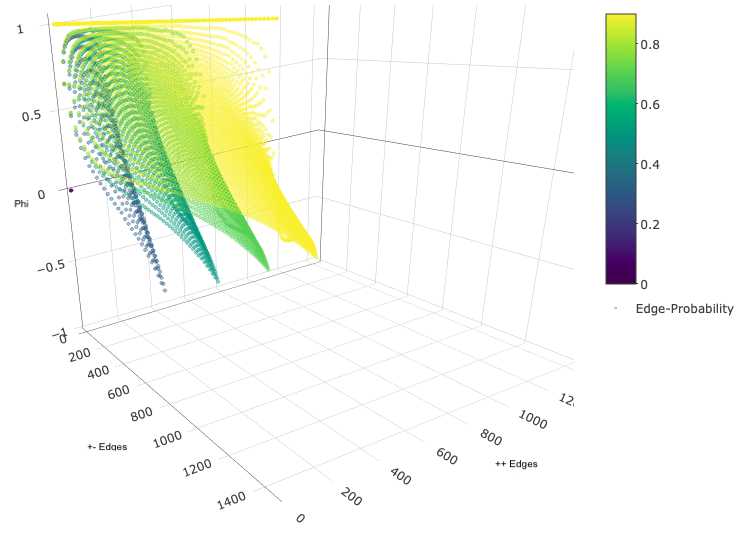}
\vspace{-6mm}
\caption{{\small Maximum $\phi$ under the Erd\H os-R\' enyyi model: Number of edges per configuration $++,+-$ vs. $\phi$ and the sampling probability. Networks in yellow indicate higher density (edge-probability). Higher MSCC is achieved for networks with higher proportion of edges labeled $++$ than $+-$. The yellow line is parallel to the '++' axis}}
\label{fig:ER-maxcorrelation}
\end{center}
\vspace{-5mm}
\end{figure}

\begin{figure*}[t]
\begin{center}
\includegraphics[width=8.72cm]{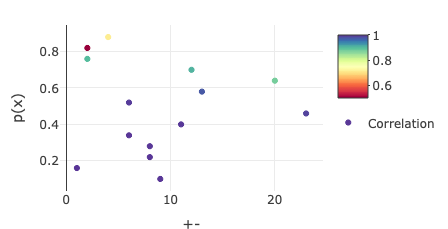}
\includegraphics[width=8.72cm]{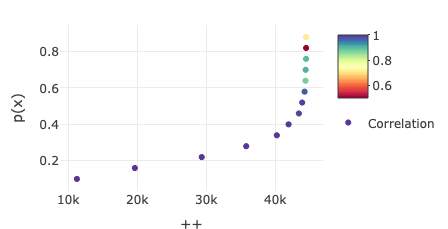}
\caption{{\small Maximum correlation for the GF model . 
}}
\vspace{-0.4cm}
\label{fig:PPP-p-vs-maxcorrelation}
\end{center}
\end{figure*}

\begin{figure*}
\begin{center}
$p_2=0.4$ \phantom{spacespacespacespacespacespacespacespacespacespace} $p_2=0.5$\\
\vspace{-1mm}
\includegraphics[width=8.745cm]{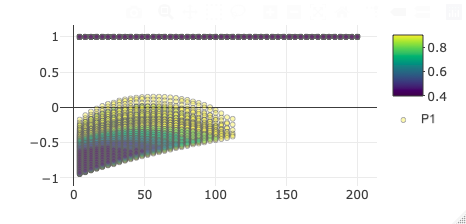}
\includegraphics[width=8.745cm]{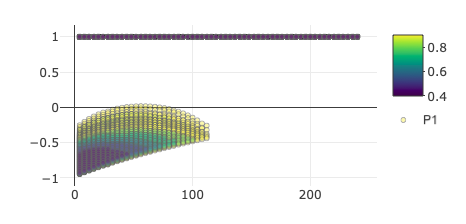}\\
%
$p_2=0.6$ \phantom{spacespacespacespacespacespacespacespacespacespace} $p_2=0.7$\\
\vspace{-1mm}
\includegraphics[width=8.745cm]{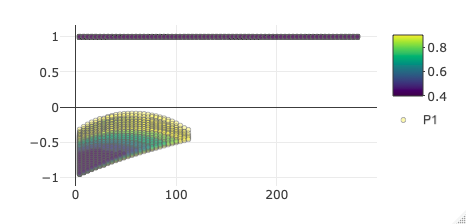}
\includegraphics[width=8.745cm]{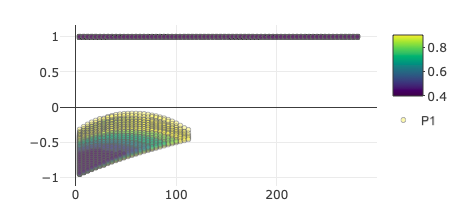}
\vspace{-0.3cm}
\caption{{\small Maximum correlation under $2\times 2$ SBM (undirected). 
Horizontal axis $=\#$ of $++$ edges.
Subplots: $p_1$ for $p_2=\{0.4,0.5,0.6,0.7\}$
}
}
\label{fig:SBM-maxcorrelation}
\end{center}
\vspace{-5mm}
\end{figure*}

In this experiment we evaluate our framework to perform goodness-of-fit based on the probabilities from Theorem \ref{thm:mainthm}. We fitted several real world datasets using the 
the Stochastic Block Model (SBM) \cite{holland1983stochastic} model for varying numbers of blocks. 
A summary of the characteristics of each dataset is presented in Table \ref{tab:data}.
\begin{table}[]
\caption{Real Network Characteristics}
\label{tab:data}
\vspace{-2mm}
\begin{tabular}
{|l|l|l|l|l|}
\hline
Name & Nodes  & Edges & Attribute (+) & Ref \\ \hline
adjnoun & 109 & 11881 & 3564 & \cite{newman2006finding} \\
dolphins & 51 & 2601 & 770 & \cite{lusseau2003bottlenose} \\
polbooks & 103 & 10609 & 3180 & \cite{booksPol} \\
football & 113 & 12769 & 3800 & \cite{girvan2002community} \\
lesmis & 74 & 5476 & 1642 & \cite{knuth1993stanford} \\
\hline
\end{tabular}
\vspace{-4mm}
\end{table}
Figure \ref{fig:modeling2} shows the result of this analysis where the horizontal axis correspond to the various choices of the number of blocks and the vertical axis, the probability of representing the reference network. As we can see there, each network has a different optimal choice of $K$. For instance, the \texttt{lesmis} dataset has optimal $K=4$. Most of the datasets (all except football) have high probability of representing the reference graph even with a small value of $K$. The results in the figure also show that larger values of $K$ are less likely to represent the target network, which is obviously the case.  
The benefit of our work is to provide the tool to identify the conditions (in terms of the model probabilities) for the representation of the correlation of graphs with node attributes with the only constraint that the model belongs to the class of models $\mathcal C$.
Ours is a non-asymptotic framework to understand the minimal size of a model (in terms of candidate edges and their probabilities) that can generate graphs with specific attribute correlations.

\subsection{Simulations and Evaluation of Models}
We evaluate empirically the values for maximum correlation that can be modeled under the 
GF and SBM models (additional experiments appear in the Appendix).

Figure \ref{fig:SBM-maxcorrelation} shows the maximum correlation as a function of the edge probability of SBM.
Notice that since in SBM the parameters is a matrix, the maximum correlation corresponds to a spectrum of values that reflect how the parameters interact. Namely, for a $2\times 2$ SBM model with parameters $\Theta=\begin{bmatrix}
p_1 & p_2 \\
p_3 & p_4 
\end{bmatrix}$ 
$p_1$ and $p_4$ has an indirect impact on the maximum correlation achievable, e.g,  values of the edge probability of $0.4$ can lead to a maximum correlation of $1$. On the other hand,  $p_2$ and $p_3$ have a more direct impact on the maximum correlation and only large values of $p_2,p_3$ can lead to a maximum correlation of $1$. This is somewhat counter-intuitive but could be explained from the point of view that a within cluster connectivity of $0.4$ may be sufficient to achieve the largest possible correlation. 

Figure \ref{fig:PPP-p-vs-maxcorrelation}, presents the results for attributed graphs sampled from the graph frequency (GF) model. For the case of the analysis of edge probabilities we used the same parameters used in \citet{cai:2016}: $\check\alpha = 0.5$, $\check\beta=1$, $\check\gamma=3$ for the three-parameter generalized beta process that defines the edge-probabilities. As in \citet{cai:2016}, we stop the process at $2000$ iterations and binarize the graphs. 
To make comparison fair, we use a number of nodes $N=1600$ and we study the effect of the likelihood among the number of $\beta_1$ and $\beta_2$ on the correlation. Notice that the monotonicity remains despite the complexity of the model. 

\begin{figure}[h]
\begin{center}
\includegraphics[width=6.5cm]{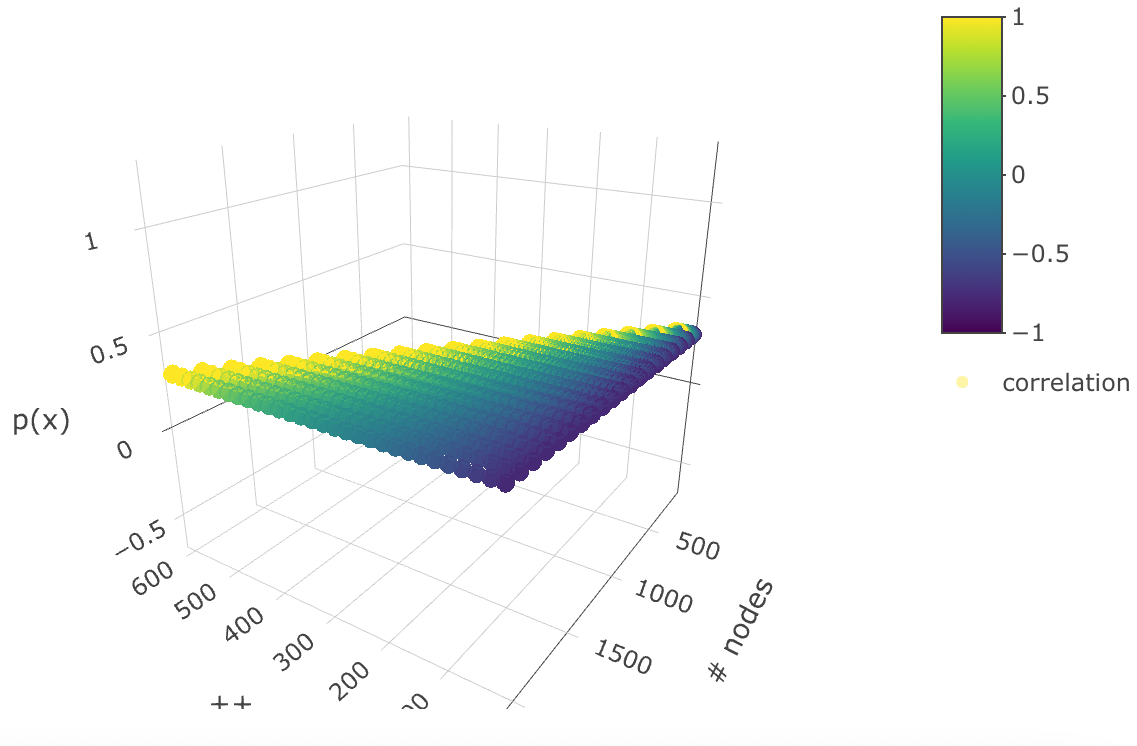}
\includegraphics[width=6.5cm]{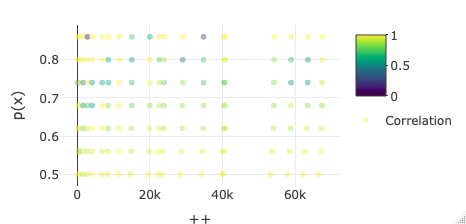}
\caption{{\small Effect of time for: left: the ER model right: GF model 
}}
\label{fig:ER-PPP-time-vs-maxcorrelation}
\end{center}
\end{figure}

Finally, Figure \ref{fig:ER-PPP-time-vs-maxcorrelation} shows the results of the effect of the number of nodes on the correlation. we explore the effect of $N$ for the GF model for the same $N$ in the range $[100,2000]$ with step size of $100$. We choose to plot a two-dimensional representation of the effect of $\beta_1$ on the maximum correlation because the maximum correlation does not define a clearly separated section of the graph space. This is due to both the complexity of the GF model and the non-trivial relation of the attribute marginal. Notice that due to the binarization the range of density of the nodes is wider than for the $ER$ model. Likewise, we show the case for the ER model with the same parameters, except for $N$, on the left sub-plot. We varied $N$ in the range $[100,2000]$ with step size of $100$. 
For the ER model we plot a three dimensional representation of the number of nodes, the number of $\beta_1$ edges, the attribute probability $p(X)$, and the effect on the maximum correlation (color-coded).

Figure \ref{fig:ER-maxcorrelation} in the Appendix shows the maximum correlation as a function of the edge probability of the ER model as an additional illustrative example. This evaluation shows two important insights obtained from our theoretical framework. First the values of the correlations along $\beta_1$ values are monotonically increasing. Second, the maximum correlation of node attributes becomes more restricted as the value of the structural parameter $\Theta_{ER}=[p(x),N]$ increases. Notice that for this experiment we varied p(x) for $N=1600$. We also studied the effect over $N$ as shown later in this section.


\vspace{-3mm}
\section{Related Work}

Prior work on models for attributed-graphs include \cite{bothorel:2015,Silva:2012,Kolar:2014}. 
Goodness-of-fit measures for graphs are a thriving area of research (\citet{pmlr-v80-yang18c,leppala2017admixturegraph,chen2019bootstrap} etc.). However, goodness-of-fit for labeled graphs is less explored (\citet{faloutsos2019social,adriaens2020block}. To the best of our knowledge this problem has not been fully addressed in other works most of which rely on traditional metrics such as $r^2$ (\citet{dimitriadis2020triage}). We define  ``characteristics'' w.r.t the parameters such that both structure and node-attributes are captured simultaneously, as opposed to separately through traditional metrics.

Our work is related to the threshold phenomena 
in random graphs (\citet{mossel2018proof,mossel2014consistency,kalai2015sharp,deshpande2018contextual}. %
The closest work to ours is \cite{mossel2018proof} which presented a solution to the clustering problem originally proposed by \citet{Decelle}, namely, the \emph{Threshold Conjecture}. However, the labels used in \cite{mossel2018proof} correspond to the block assignment of SBM - thus a clustering problem pertaining to the graph structure. Unlike this problem, ours considers labels drawn from a Bernoulli distribution and may define highly non-symmetric states that are fitted for the marginal distribution of node attributes that have little or no relation to the block community structure - thus ours 
is a sampling problem. 
Earlier threshold results for Boolean functions in graphs with  symmetry, influence, and pivotality were reported by \cite{kalai2006perspectives}. %

The mean square contingency coefficient or $\phi$-coefficient is a measure of association between two binary variables. \citet{davenport51sanhury}\cite{Davenport91} 
proved the maximum values for $\phi$ in the case of a constant contingency table. However, this is not directly applicable to our analysis because the table in our case comes from random graphs. Thus, ours is a generalization of this work for the case of tables derived from random graphs.
%

 \vspace{-2mm}
\section{Discussion and Conclusion}
\vspace{-1mm}
In this \paper, we presented both sampling guarantees  
of a general class $\mc C$ of probabilistic generative models and a framework for sampling graph 
structure and node-attributes. Specifically, we introduced: 
%
%
the maximal marginal-error associated with the structural and attribute margins of the 
model and,
%
an information-theoretical and probabilistic guarantee for a general class of models $\mc C$ equivalent to a possibly sparse parametric matrix. 
%
We also provided examples of the applicability of the analysis and 
an example of the probability of sample graphs (with specific auto-correlation) vs. the size of model in terms of its candidate edges. 
Our framework is 
focused on the assumption of sampling-agnostic exchangeability of structure and exchangeability preservation.

The main challenge we aimed to solve was assessing the correlation preservation of a model because preserving 
structure and attribute distribution can be done with existing Method of Moments and other statistical tools. 
Extensions to multiple-labeled graphs is not straightforward because the thresholds of each specific family of distributions may be considered. %

Our work facilitates an understanding of characteristics of a generative model of node-attributed graphs and can be applied to hypothesis testing. %
It seeks to understand what type of data can be represented with a model using our probabilistic analysis. It can be used to reduce computational costs for model selection, network hypothesis testing, and among other possible applications~\cite{Asta2015}.

Identifying theoretical constraints in probabilistic models of networks is relevant to the machine learning community because a thorough understanding of representation constraints in random graphs can help research communities determine which models are usable, for instance via hypothesis test – this is highly relevant to such varied domains as relational learning, collaborative filtering, graph mining, etc., where evaluation of graph models are useful.


\vspace{-1mm}
\section*{Acknowledgements}

This work is partially supported by NSF III 2046795, IIS 1909577, CCF 1934986, NIH 1R01MH116226-01A, NIFA award 2020-67021-32799, the Alfred P. Sloan Foundation, Google Inc, and by a Future Faculty Fellowship from the Computer Science Department at the University of Illinois at Urbana-Champaign.

\bibliographystyle{plainnat}

\small{\bibliography{probles-all}}

\begin{thebibliography}{44}
\providecommand{\natexlab}[1]{#1}
\providecommand{\url}[1]{\texttt{#1}}
\expandafter\ifx\csname urlstyle\endcsname\relax
  \providecommand{\doi}[1]{doi: #1}\else
  \providecommand{\doi}{doi: \begingroup \urlstyle{rm}\Url}\fi

\bibitem[boo()]{booksPol}
Social organizational network analysis software.
\newblock URL \url{http://www.orgnet.com/}.

\bibitem[Adriaens et~al.(2020)Adriaens, Mara, Lijffijt, and
  De~Bie]{adriaens2020block}
Florian Adriaens, Alexandru Mara, Jefrey Lijffijt, and Tijl De~Bie.
\newblock Block-approximated exponential random graphs.
\newblock In \emph{2020 IEEE 7th International Conference on Data Science and
  Advanced Analytics (DSAA)}, pages 70--80. IEEE, 2020.

\bibitem[Ahmed and Xing(2009)]{Ahmed:09}
Amr Ahmed and Eric~P. Xing.
\newblock Recovering time-varying networks of dependencies in social and
  biological studies.
\newblock \emph{Proceedings of the National Academy of Sciences}, 106\penalty0
  (29):\penalty0 11878--11883, 2009.

\bibitem[Asta and Shalizi(2015)]{Asta2015}
Dena~Marie Asta and Cosma~Rohilla Shalizi.
\newblock Geometric network comparisons.
\newblock In \emph{UAI}, 2015.

\bibitem[Bansal et~al.(2004)Bansal, Blum, and Chawla]{Bansal:2004}
Nikhil Bansal, Avrim Blum, and Shuchi Chawla.
\newblock Correlation clustering.
\newblock \emph{Mach. Learn.}, 56\penalty0 (1-3):\penalty0 89--113, June 2004.
\newblock ISSN 0885-6125.
\newblock \doi{10.1023/B:MACH.0000033116.57574.95}.
\newblock URL \url{https://doi.org/10.1023/B:MACH.0000033116.57574.95}.

\bibitem[Barabasi and Albert(1999)]{BarAlb99}
A.~Barabasi and R.~Albert.
\newblock Emergence of scaling in random networks.
\newblock \emph{Science}, 286:\penalty0 509--512, 1999.

\bibitem[Bothorel et~al.(2015)Bothorel, Cruz, Magnani, and
  Micenkova]{bothorel:2015}
Cecile Bothorel, Juan~David Cruz, Matteo Magnani, and Barbora Micenkova.
\newblock Clustering attributed graphs: Models, measures and methods.
\newblock \emph{Network Science}, 3\penalty0 (3):\penalty0 408–444, 2015.

\bibitem[Cai et~al.(2016)Cai, Campbell, and Broderick]{cai:2016}
Diana Cai, Trevor Campbell, and Tamara Broderick.
\newblock Edge-exchangeable graphs and sparsity.
\newblock In D.~D. Lee, M.~Sugiyama, U.~V. Luxburg, I.~Guyon, and R.~Garnett,
  editors, \emph{Advances in Neural Information Processing Systems 29}, pages
  4249--4257. Curran Associates, Inc., 2016.

\bibitem[Chen and Onnela(2019)]{chen2019bootstrap}
Sixing Chen and Jukka-Pekka Onnela.
\newblock A bootstrap method for goodness of fit and model selection with a
  single observed network.
\newblock \emph{Scientific reports}, 9\penalty0 (1):\penalty0 1--12, 2019.

\bibitem[Cramer(1946)]{cramer1946mathematical}
Harold Cramer.
\newblock Mathematical methods of statistics, princeton univ.
\newblock \emph{Press, Princeton, NJ}, 1946.

\bibitem[Crane and Dempsey(2016)]{crane2016edge}
Harry Crane and Walter Dempsey.
\newblock Edge exchangeable models for network data.
\newblock \emph{arXiv preprint arXiv:1603.04571}, 2016.

\bibitem[de~Solla~Price(1976)]{de1976general}
Derek de~Solla~Price.
\newblock A general theory of bibliometric and other cumulative advantage
  processes.
\newblock \emph{Journal of the Association for Information Science and
  Technology}, 27\penalty0 (5):\penalty0 292--306, 1976.

\bibitem[Decelle et~al.(2011)Decelle, Krzakala, Moore, and
  Zdeborov\'a]{Decelle}
Aurelien Decelle, Florent Krzakala, Cristopher Moore, and Lenka Zdeborov\'a.
\newblock Asymptotic analysis of the stochastic block model for modular
  networks and its algorithmic applications.
\newblock \emph{Phys. Rev. E}, 84:\penalty0 066106, Dec 2011.

\bibitem[Deshpande et~al.(2018)Deshpande, Sen, Montanari, and
  Mossel]{deshpande2018contextual}
Yash Deshpande, Subhabrata Sen, Andrea Montanari, and Elchanan Mossel.
\newblock Contextual stochastic block models.
\newblock \emph{Advances in Neural Information Processing Systems}, 31, 2018.

\bibitem[Dimitriadis et~al.(2020)Dimitriadis, Poiitis, Faloutsos, and
  Vakali]{dimitriadis2020triage}
Ilias Dimitriadis, Marinos Poiitis, Christos Faloutsos, and Athena Vakali.
\newblock Triage: Temporal twitter attribute graph patterns.
\newblock In \emph{Proceedings of the 10th International Conference on Web
  Intelligence, Mining and Semantics}, pages 44--53, 2020.

\bibitem[El-Sanhury and Davenport()]{davenport51sanhury}
N.~El-Sanhury and E~Davenport.
\newblock Phi/phimax: Review and synthesis educational and psychological
  measurement.
\newblock \emph{Educational and Psychological Measurement}, 51\penalty0
  (4):\penalty0 821--828.

\bibitem[Ernest C.~Davenport and El-Sanhurry(1991)]{Davenport91}
Jr. Ernest C.~Davenport and Nader~A. El-Sanhurry.
\newblock Phi/phimax: Review and synthesis.
\newblock \emph{Educational and Psychological Measurement}, 51\penalty0
  (4):\penalty0 821--828, 1991.

\bibitem[Eswaran et~al.(2019)Eswaran, Rabbany, Dubrawski, and
  Faloutsos]{faloutsos2019social}
Dhivya Eswaran, Reihaneh Rabbany, Arthur Dubrawski, and Christos Faloutsos.
\newblock Social-affiliation networks: Patterns and the soar model.
\newblock In \emph{Machine Learning and Knowledge Discovery in Databases:
  European Conference, ECML PKDD 2018, Dublin, Ireland, September 10--14, 2018,
  Proceedings, Part II}, volume 11052, page 105. Springer, 2019.

\bibitem[Girvan and Newman(2002)]{girvan2002community}
Michelle Girvan and Mark~EJ Newman.
\newblock Community structure in social and biological networks.
\newblock \emph{Proceedings of the national academy of sciences}, 99\penalty0
  (12):\penalty0 7821--7826, 2002.

\bibitem[Holland et~al.(1983)Holland, Laskey, and
  Leinhardt]{holland1983stochastic}
Paul~W Holland, Kathryn~Blackmond Laskey, and Samuel Leinhardt.
\newblock Stochastic blockmodels: First steps.
\newblock \emph{Social networks}, 5\penalty0 (2):\penalty0 109--137, 1983.

\bibitem[Holme and Kim(2002)]{holme2002growing}
Petter Holme and Beom~Jun Kim.
\newblock Growing scale-free networks with tunable clustering.
\newblock \emph{Physical review E}, 65\penalty0 (2):\penalty0 026107, 2002.

\bibitem[Jeong et~al.(2000)Jeong, Tombor, Albert, Oltvai, and
  Barabasi]{Jeong:00}
H.~Jeong, B.~Tombor, R.~Albert, Z.~N. Oltvai, and A.~L. Barabasi.
\newblock The large-scale organization of metabolic networks.
\newblock \emph{Nature}, 407\penalty0 (6804):\penalty0 651--654, 10 2000.

\bibitem[Kalai and Mossel(2015)]{kalai2015sharp}
Gil Kalai and Elchanan Mossel.
\newblock Sharp thresholds for monotone non-boolean functions and social choice
  theory.
\newblock \emph{Mathematics of Operations Research}, 40\penalty0 (4):\penalty0
  915--925, 2015.

\bibitem[Kalai and Safra(2006)]{kalai2006perspectives}
Gil Kalai and Shmuel Safra.
\newblock Perspectives from mathematics, computer science, and economics.
\newblock \emph{Computational complexity and statistical physics}, page~25,
  2006.

\bibitem[Kim and Leskovec(2010)]{Myunghwan:10}
Myunghwan Kim and Jure Leskovec.
\newblock Multiplicative attribute graph model of real-world networks.
\newblock In \emph{Algorithms and Models for the Web-Graph}, volume 6516 of
  \emph{Lecture Notes in Computer Science}, pages 62--73, 2010.
\newblock ISBN 978-3-642-18008-8.

\bibitem[Knuth(1993)]{knuth1993stanford}
Donald~E Knuth.
\newblock \emph{The Stanford GraphBase: a platform for combinatorial
  computing}.
\newblock Addison-Wesley, 1993.

\bibitem[Kolar et~al.(2014)Kolar, Liu, and Xing]{Kolar:2014}
Mladen Kolar, Han Liu, and Eric~P. Xing.
\newblock Graph estimation from multi-attribute data.
\newblock \emph{J. Mach. Learn. Res.}, 15\penalty0 (1):\penalty0 1713--1750,
  January 2014.
\newblock ISSN 1532-4435.

\bibitem[Lepp{\"a}l{\"a} et~al.(2017)Lepp{\"a}l{\"a}, Nielsen, and
  Mailund]{leppala2017admixturegraph}
Kalle Lepp{\"a}l{\"a}, Svend~V Nielsen, and Thomas Mailund.
\newblock admixturegraph: an r package for admixture graph manipulation and
  fitting.
\newblock \emph{Bioinformatics}, 33\penalty0 (11):\penalty0 1738--1740, 2017.

\bibitem[Lusseau et~al.(2003)Lusseau, Schneider, Boisseau, Haase, Slooten, and
  Dawson]{lusseau2003bottlenose}
David Lusseau, Karsten Schneider, Oliver~J Boisseau, Patti Haase, Elisabeth
  Slooten, and Steve~M Dawson.
\newblock The bottlenose dolphin community of doubtful sound features a large
  proportion of long-lasting associations.
\newblock \emph{Behavioral Ecology and Sociobiology}, 54\penalty0 (4):\penalty0
  396--405, 2003.

\bibitem[Mossel et~al.(2014)Mossel, Neeman, and Sly]{mossel2014consistency}
Elchanan Mossel, Joe Neeman, and Allan Sly.
\newblock Consistency thresholds for binary symmetric block models.
\newblock \emph{arXiv preprint arXiv:1407.1591}, 2014.

\bibitem[Mossel et~al.(2018)Mossel, Neeman, and Sly]{mossel2018proof}
Elchanan Mossel, Joe Neeman, and Allan Sly.
\newblock A proof of the block model threshold conjecture.
\newblock \emph{Combinatorica}, 38\penalty0 (3):\penalty0 665--708, 2018.

\bibitem[Newman(2006)]{newman2006finding}
Mark~EJ Newman.
\newblock Finding community structure in networks using the eigenvectors of
  matrices.
\newblock \emph{Physical review E}, 74\penalty0 (3):\penalty0 036104, 2006.

\bibitem[Pfeiffer~III et~al.(2014)Pfeiffer~III, Moreno, La~Fond, Neville, and
  Gallagher]{Pfeiffer:14}
J.~J. Pfeiffer~III, S.~Moreno, T.~La~Fond, J.~Neville, and B.~Gallagher.
\newblock Attributed graph models: Modeling network structure with correlated
  attributes.
\newblock In \emph{Proceedings of the Twenty-Third International Conference on
  World Wide Web}, WWW '14, pages 831--842, 2014.
\newblock ISBN 978-1-4503-2744-2.

\bibitem[Powers(2020)]{powers2020evaluation}
David~MW Powers.
\newblock Evaluation: from precision, recall and f-measure to roc,
  informedness, markedness and correlation.
\newblock \emph{arXiv preprint arXiv:2010.16061}, 2020.

\bibitem[Psorakis et~al.(2012)Psorakis, Roberts, Rezek, and
  Sheldon]{Psorakis:12}
Ioannis Psorakis, Stephen~J. Roberts, Iead Rezek, and Ben~C. Sheldon.
\newblock Inferring social network structure in ecological systems from
  spatio-temporal data streams.
\newblock \emph{Journal of The Royal Society Interface}, 2012.
\newblock ISSN 1742-5689.

\bibitem[R\'enyi and Erd\H{o}s(1959)]{renyi1959random}
A~R\'enyi and P~Erd\H{o}s.
\newblock On random graph.
\newblock \emph{Publicationes Mathematicate}, 6:\penalty0 290--297, 1959.

\bibitem[Shalizi and Rinaldo(2013)]{shalizi:2013}
Cosma~Rohilla Shalizi and Alessandro Rinaldo.
\newblock Consistency under sampling of exponential random graph models.
\newblock \emph{Ann. Statist.}, 41\penalty0 (2):\penalty0 508--535, 04 2013.

\bibitem[Shore and Lubin(2015)]{SHORE201516}
Jesse Shore and Benjamin Lubin.
\newblock Spectral goodness of fit for network models.
\newblock \emph{Social Networks}, 43:\penalty0 16--27, 2015.
\newblock ISSN 0378-8733.

\bibitem[Silva et~al.(2012)Silva, Meira, and Zaki]{Silva:2012}
Arlei Silva, Wagner Meira, Jr., and Mohammed~J. Zaki.
\newblock Mining attribute-structure correlated patterns in large attributed
  graphs.
\newblock \emph{Proc. VLDB Endow.}, 5\penalty0 (5):\penalty0 466--477, January
  2012.
\newblock ISSN 2150-8097.

\bibitem[Spyropoulou et~al.(2014)Spyropoulou, De~Bie, and
  Boley]{Spyropoulou:14}
Eirini Spyropoulou, Tijl De~Bie, and Mario Boley.
\newblock Interesting pattern mining in multi-relational data.
\newblock \emph{Data Mining and Knowledge Discovery}, 28\penalty0 (3):\penalty0
  808--849, 2014.
\newblock ISSN 1384-5810.

\bibitem[Weckbecker et~al.(2022)Weckbecker, Xu, and
  Reinert]{weckbecker2022rkhs}
Moritz Weckbecker, Wenkai Xu, and Gesine Reinert.
\newblock On rkhs choices for assessing graph generators via kernel stein
  statistics.
\newblock \emph{arXiv preprint arXiv:2210.05746}, 2022.

\bibitem[Yang et~al.(2018)Yang, Liu, Rao, and Neville]{pmlr-v80-yang18c}
Jiasen Yang, Qiang Liu, Vinayak Rao, and Jennifer Neville.
\newblock Goodness-of-fit testing for discrete distributions via stein
  discrepancy.
\newblock In Jennifer Dy and Andreas Krause, editors, \emph{Proceedings of the
  35th International Conference on Machine Learning}, volume~80 of
  \emph{Proceedings of Machine Learning Research}, pages 5561--5570, 2018.

\bibitem[Zhou and Nakhleh(2011)]{Zhou:11}
Wanding Zhou and Luay Nakhleh.
\newblock Properties of metabolic graphs: Biological organization or
  representation artifacts?
\newblock \emph{BMC Bioinformatics}, 12\penalty0 (1):\penalty0 1--12, 2011.
\newblock ISSN 1471-2105.

\bibitem[Zhu(2020)]{zhu2020performance}
Qiuming Zhu.
\newblock On the performance of matthews correlation coefficient (mcc) for
  imbalanced dataset.
\newblock \emph{Pattern Recognition Letters}, 136:\penalty0 71--80, 2020.

\end{thebibliography}


\newpage
\onecolumn
\section{Appendix}

\subsection{Additional Details of the Proofs of the Theorems}
\label{sec:appdx-proofs}

\begin{proof}{\emph{Lemma \ref{thm:corr-bound}}} %
Since the unique probabilities $\pi_j\in \mathbf U$ are Bernoulli-distributed, the number of edges to sample $y_j\sim Bin(n_j,\pi_j)$ are binomial-distributed. 
Consider the tail bounds of the binomial distribution (Arriata and Gordon, 1989) 
:
\vspace{-0.5em}
\[
\begin{array}{l}
Pr(X\leq k;n,p)\leq e^{\left(-n D_{KL}\left(\frac{k}{n}||p\right)\right)} \mbox{ if } 0<\frac{k}{n}<p;\\
Pr(X\geq k;n,p)\leq e^{\left(-n D_{KL}\left(\frac{k}{n}||p\right)\right)} \mbox{ if } p<\frac{k}{n}<1.\\
\end{array}
\vspace{-0.5em}
\]
Then,
\[
\footnotesize{
\begin{array}{l}
Pr(y_j\leq N_{1j}+N_{3_j};n_j,\pi_j)\leq e^{\left(-n_j D_{KL}\left(r_j||\pi_j\right)\right)} \mbox{ if } 0<r_j<\pi_j;\\
Pr(y_j\geq N_{1j}+N_{3_j};n_j,\pi_j)\leq e^{\left(-n_j D_{KL}\left(r_j||\pi_j\right)\right)} \mbox{ if } \pi_j<r_j<1.\\
\end{array}
}
\]
To find $\exp(-w)<10^{-10}$ consider $-w<-10\log{10}\Rightarrow w> 23.03$.
Then
\[
\left\{\begin{array}{l}
Pr(y_j> N_{1j}+N_{3_j};n_j,\pi_j)> 1-10^{-10}, \mbox { if } \left(n_j D_{KL}\left(r_j||\pi_j\right)\right)>23.03 \mbox{ and } 0<r_j<\pi_j;\\
Pr(y_j< N_{1j}+N_{3_j};n_j,\pi_j)> 1-10^{-10},\mbox { if } \left(n_j D_{KL}\left(r_j||\pi_j\right)\right)>23.03 \mbox{ and } \pi_j<r_j<1. \\
\end{array}\right.
\vspace{-1.7em}
\]
\end{proof}

\begin{proof} {\em Lemma \ref{thm:monincreasing}}
Consider the function $\rho:\mc U\rightarrow [-1,1]$ 
defined in Equation~\eqref{eq:rho_1}.
This function is monotonically increasing along $\beta_1$ because
\[\frac{\partial \rho}{\partial \beta_1}=\frac{4(\beta_1+1)\beta_3+2(\beta_1-1)^2-6\beta_3^2}{((\beta_1-\beta_3)^2-1)^2}  \geq 0.\]
Likewise, this function is monotonically increasing along $\beta_3$ because
\[\frac{\partial \rho}{\partial \beta_3}=\frac{4(\beta_3+1)\beta_1+2(\beta_3-1)^2-6\beta_1^2}{((\beta_3-\beta_1)^2-1)^2}  \geq 0.\]
Then, by Remark \ref{rem:realbimonotonic}, after considering the dimensions $x_1,x_2$ as $\beta_1, \beta_3$, it follows that for $\vec\beta\preceq\vec\beta'\Rightarrow \rho(\vec\beta)\leq\rho(\vec\beta')$.
\end{proof}

\begin{proof} {\em Lemma \ref{thm:maxepsilon}}\\
The conditions
$|\beta_{1}-\tilde\beta_{1}|<\epsilon_1 \mbox{  and  } |\beta_{3}-\tilde\beta_{3}|<\epsilon_3$
are equivalent to 
$\beta_{1}-\epsilon_1<\tilde\beta_{1}<\beta_{1}+\epsilon_1 \mbox{  and  } \beta_{3}-\epsilon_3<\tilde\beta_{3}<\beta_{3}+\epsilon_3$.
Thus, the solution is defined by the values $\tilde\beta_1,\tilde\beta_3$ that maximize the difference of the correlations in the squared region
$(\beta_1,\beta_3)+[-\epsilon_1,\epsilon_1]\times [-\epsilon_3,\epsilon_3]$.

From Lemma \ref{thm:monincreasing}, we know that the correlation is monotonically increasing with respect to projections in each dimension $\beta_1,\beta_3$. Then
\begin{align*}
    \epsilon&=\max(|\rho_{IN}-\rho_{OUT}|)=\max(|\rho(\tilde\beta_1,\tilde\beta_3)-\rho(\beta_1,\beta_3)|)\\&=\max(\rho(\beta_1+\epsilon_1,\beta_3+\epsilon_3)-\rho(\beta_1,\beta_3),-\rho(\beta_1-\epsilon_1,\beta_3-\epsilon_3)+\rho(\beta_1,\beta_3)).\\
\end{align*}
Thus, for values of $\rho(\tilde\beta_1,\tilde\beta_3)$ that oversample $\rho_{IN}$ we can use the expression:
\begin{align*}
    \rho(\beta_1+\epsilon_1,\beta_3+\epsilon_3)&-\rho(\beta_1,\beta_3)
    \\&=\frac{(\beta_3+\epsilon_3)-\left(\beta_3+\epsilon_3+\frac{1-\beta_1-\epsilon_1-\beta_3-\epsilon_3}{2}\right)^2}{(\beta_3+\epsilon_3+(1-\beta_1-\epsilon_1-\beta_3-\epsilon_3)/2)(1-(\beta_3+\epsilon_3+(1-\beta_1-\epsilon_1-\beta_3-\epsilon_3))/2)}\\
    &\quad-\frac{(\beta_3)-\left(\beta_3+\frac{1-\beta_1-\beta_3}{2}\right)^2}{(\beta_3+(1-\beta_1-\beta_3)/2)(1-(\beta_3+(1-\beta_1-\beta_3))/2)}.
\end{align*}
Therefore,
\begin{align*}
    \epsilon&=
\frac{\beta_3-(p+\Delta p)^2}{(p+\Delta p)(1-p-\Delta p)}
-
\frac{\beta_3-p^2}{p(1-p)},
\end{align*}
where $\Delta p=\epsilon_3+\frac{-\epsilon_1-\epsilon_3}{2}$.

\end{proof}

\begin{proof} {\em Theorem \ref{thm:mainthm}}

Consider the case of a tight bound on $\vec\beta$ and let $\beta_i$ be each entry of $\vec\beta$ (i.e., derived from the data) and let $\tilde\beta_i$ be associated to the output/sampled graph.
By lemma \ref{thm:maxepsilon} there is an $\epsilon$:
\[|\beta_{1}-\tilde\beta_{1}|<\epsilon_1 \mbox{  and  } |\beta_{2}-\tilde\beta_{2}|<\epsilon_2 \Rightarrow |\rho_{IN}-\rho_{OUT}|<\epsilon,\]

Consider $\beta_i$ constant. Replacing the definitions of the ratios in 
$|\beta_{i}-\tilde\beta_{i}|<\epsilon_i$ gives us
$\left|\beta_{i}-\frac{\sum_{j}^{\kappa} \chi_{ij}}{\sum_j^{\kappa} y_j}\right|<\epsilon_i$.

Notice that $\chi$ and $y_j$ are not independent and $\chi_{ij}<y_j$. In fact, $\chi_{ij}<N_{ij}$.
Now the value of $\chi$ can be deterministic or probabilistic. %
Consider the deterministic case (we will consider the probabilistic case in Thm \ref{thm:mainthmrnd}):
\[\frac{\sum_{j}^{\kappa} \chi_{ij}}{\beta_{i}+\epsilon_i}<\sum_j^{\kappa} y_j<\frac{\sum_{j}^{\kappa} \chi_{ij}}{\beta_{i}-\epsilon_i},
\]
or
\[
Pr\left(\sum_j^{\kappa} y_j<\frac{\sum_{j}^{\kappa} \chi_{ij}}{\beta_{i}-\epsilon_i}\right)
-
Pr\left(\sum_j^{\kappa} y_j<\frac{\sum_{j}^{\kappa} \chi_{ij}}{\beta_{i}+\epsilon_i}\right).\]
%
Let $X=\sum_j^{\kappa} y_j$,
then $Pr\left(X<a\right)
-
Pr\left(X<b\right)>1-\delta$ can be written as
$\left(Pr\left(X>a\right)
+
Pr\left(X<b\right)\right)<\delta
$. 
%
Alternately, considering $Pr\left(X>a\right)
<\delta_1$
and
$
Pr\left(X<b\right)<\delta_2$, then
\[\left(Pr\left(X>a\right)
+
Pr\left(X<b\right)\right)<\delta_1+\delta_2.\]
Consider $a=(1+\xi)\mu$ for $\mu=E[X]$ and $0<\xi<1$, then by multiplicative Chernoff bound:
$$Pr\left(X>a\right)\leq e^{-\frac{\xi^2\mu}{3}}=e^{-\frac{{({a-\mu})}^2}{3\mu}},$$
where the equality follows by $\xi=\frac{a-\mu}{\mu}$.
Consider $b=(1-\xi')\mu$ for $\mu=E[X]$ and $0<\xi'<1$, then, by the same bound:
$$Pr\left(X<b\right)\leq e^{-\frac{\xi'^2\mu}{2}}=e^{-\frac{{({\mu}-b)}^2}{2\mu}},$$
 --- recall $\left(Pr\left(X>a\right)
+
Pr\left(X<b\right)\right)<\delta_1+\delta_2$ -- Then:
$Pr(|\beta_{i}-\tilde\beta_{i}|<\epsilon_i)>1-(e^{-\frac{{({a-\mu})}^2}{3\mu}}+e^{-\frac{{({\mu}-b)}^2}{2\mu}})
$ or equivalently:
$$
Pr(|\beta_{i}-\tilde\beta_{i}|<\epsilon_i)>1-\left(e^{-\frac{{\left({\frac{\sum_{j}^{\kappa} \chi_{ij}}{\beta_{i}-\epsilon_i}-\mu}\right)}^2}{3\mu}}+e^{-\frac{{\left({\mu}-\frac{\sum_{j}^{\kappa} \chi_{ij}}{\beta_{i}+\epsilon_i}\right)}^2}{2\mu}}\right)
.$$
%
%
%
%
Thus, $P(|\rho_{IN}-\rho_{OUT}|<\epsilon)>1-\delta$ for $\delta=\sum_{i=1}^{2} \tau_i - \prod_{i=1}^{2} \tau_i$ and 
$\tau_i = \exp{\left(-{{\left({\frac{\sum_{j}^{\kappa} \chi_{ij}}{\beta_{i}-\epsilon_i}-\mu}\right)}^2}/{3\mu}\right)}+\exp{\left(-{{\left({\mu}-\frac{\sum_{j}^{\kappa} \chi_{ij}}{\beta_{i}+\epsilon_i}\right)}^2}/{2\mu}\right)}$.
\end{proof}

\begin{theorem}[Correlation Recovery - Random  $\rho_{IN}$]
\label{thm:mainthmrnd}
Let $\chi_{ij}$ be the number of edges sampled by $\mc S$ per edge-label $\psi_i$ and parameter $\pi_j$
and $\mu=\sum_j^{\kappa} y_j$ for $y_j\sim Bin(n_j,\pi_j)$ (number of edges to sample).
Then, for
any $\mc M$ and $\mc S$ and small $\delta,\epsilon_1,\epsilon_3>0$, 
the bound of the difference between the target correlation $\rho_{IN}$ and the correlation of the sampled graph $\rho_{OUT}$, as per the {limited range} 
of edges sampled $\chi_{ij}$ has probability 
\begin{equation}
P(|\rho_{IN}-\rho_{OUT}|<\epsilon)>1-\delta,
\end{equation}
for $\delta=\sum_{i=1}^{2} \tau_i + \prod_{i=1}^{2} \tau_i$,\\
where $\tau_i = \exp{\left(-{{\left({\mathbbm E\left[(\beta_{i}-\epsilon_i)^{-1}{\sum_{j}^{\kappa} \chi_{ij}}\right]-\mu}\right)}^2}/{3\mu}\right)}+\exp{\left(-{{\left({\mu}-\mathbbm E\left[(\beta_{i}+\epsilon_i)^{-1}{\sum_{j}^{\kappa} \chi_{ij}}\right]\right)}^2}/{2\mu}\right)}$\\
and $\epsilon=
\frac{\beta_3-(p+\Delta p)^2}{(p+\Delta p)(1-p-\Delta p)}
-
\frac{\beta_3-p^2}{p(1-p)}$, 
where $p=\beta_3+\frac{1-\beta_1-\beta_3}{2}$ and $\Delta p=\epsilon_3+\frac{-\epsilon_1-\epsilon_3}{2}$.
\end{theorem}

Theorem \ref{thm:mainthmrnd} describes the number of edges per parameter and edge-label samples required to maximize the probability of obtaining a target autocorrelation distribution. 

Notice that $\rho$ in this theorem is not assumed to be a constant but a random variable with a distribution. However, sampling of edges is still done by conditioning on attributes and then defining $\mb U$, $N,\psi,\beta$ and indices $\mc T$. %
This value is similar to the one obtained for Theorem \ref{thm:mainthm}, except that the value in Theorem \ref{thm:mainthmrnd} is in expectation and is only valid for values of the variables that are concave around $\mu$ (\emph{limited range} of $\chi_{ij}$).
This condition does not affect the generality of the theorem since the proof includes the relations in terms of the sampling distributions (Further details in the Appendix)  
required to maximize the probability of obtaining 
a target autocorrelation distribution. $P(U)$) 
that describe the behavior of the 
correlation.
%


\begin{proof}{\emph{Theorem \ref{thm:mainthmrnd}}}

The following is the full proof. 
As in the previous case, consider the case of a tight bound on $\vec\beta$ and let $\beta_i$ be each entry of $\vec\beta$ (i.e., derived from the data) and let $\tilde\beta_i$ be associated to the output/sampled graph.
\[|\beta_{1}-\tilde\beta_{1}|<\epsilon_1 \mbox{  and  } |\beta_{2}-\tilde\beta_{2}|<\epsilon_2 \Rightarrow |\rho_{IN}-\rho_{OUT}|<\epsilon.\]

Consider $\beta_i$ to be random and
%
 recall:

\[Pr\left(\sum_j^{\kappa} y_j<\frac{\sum_{j}^{\kappa} \chi_{ij}}{\beta_{i}-\epsilon_i}\right)
-
Pr\left(\sum_j^{\kappa} y_j<\frac{\sum_{j}^{\kappa} \chi_{ij}}{\beta_{i}+\epsilon_i}\right).\]
Let 
$X=\sum_j^{\kappa} y_j$, $U=\frac{\sum_{j}^{\kappa} \chi_{ij}}{\beta_{i}-\epsilon_i}$, and $V=\frac{\sum_{j}^{\kappa} \chi_{ij}}{\beta_{i}+\epsilon_i}$,
we can rewrite the above as
\begin{align*}
    &Pr\left(X<U\right)-
Pr\left(X<V\right);\\
    &Pr\left(X<U\right)=\sum_{u}Pr(X<U|U=u)Pr(U=u).
\end{align*}

Since
$$Pr\left(X<b\right)\leq e^{-\frac{\xi'^2\mu}{2}}=e^{-\frac{{({\mu}-b)}^2}{2\mu}},$$%
and
$$Pr\left(X>a\right)\leq e^{-\frac{\xi^2\mu}{3}}=e^{-\frac{{({a-\mu})}^2}{3\mu}},$$%
we have
$$Pr\left(X<V\right)\leq \sum_{v}e^{-\frac{{({\mu}-v)}^2}{2\mu}}Pr(V=v);$$
$$Pr\left(X>U\right)\leq \sum_{u}e^{-\frac{{({u-\mu})}^2}{3\mu}}Pr(U=u).$$%
Notice that $U$ and $V$ have inverse distributions: 
$$Pr\left(X<V\right)\leq \mathbbm E_{V}\left[ e^{-\frac{{({\mu}-V)}^2}{2\mu}}\right];$$
$$Pr\left(X>U\right)\leq \mathbbm E_{U}\left[e^{-\frac{{({U-\mu})}^2}{3\mu}}\right].$$
Since Jensen’s inequality is applicable on arbitrary intervals of a partially convex function as long as the function is Borel measurable, we apply the inequality to the convex region of the Gaussian density 
defined %
in $\mu \pm \sigma$ (can be determined via inflection points criteria).
In a few words, the functions above are concave around $\mu$. 
Thus, by Jensen's inequality:
$$Pr\left(X<V\right)<  e^{-\frac{{({\mu}-\mathbbm E[{V}])}^2}{2\mu}};$$
$$Pr\left(X>U\right)<  e^{-\frac{{({\mathbbm E[{U}]-\mu})}^2}{3\mu}}.$$
Following the same steps than the previous theorem, it is easy to see that
$$P(|\rho_{IN}-\rho_{OUT}|<\epsilon)>1-\sum_{i=1}^{2} \tau_i + \prod_{i=1}^{2} \tau_i,$$
where $\tau_i = \exp{\left(-{{\left({\mathbbm E\left[\frac{\sum_{j}^{\kappa} \chi_{ij}}{\beta_{i}-\epsilon_i}\right]-\mu}\right)}^2}/{3\mu}\right)}+\exp{\left(-{{\left({\mu}-\mathbbm E\left[\frac{\sum_{j}^{\kappa} \chi_{ij}}{\beta_{i}+\epsilon_i}\right]\right)}^2}/{2\mu}\right)}$\\
and $\epsilon=
\frac{\beta_1\beta_3-\gamma^2}{(\beta_1+\gamma)(\beta_3+\gamma)}-\frac{\beta_1^{'}\beta_3^{'}-{\gamma^2}^{'}}{(\beta_1^{'}+\gamma^{'})(\beta_3^{'}+\gamma^{'})}$, where $\gamma=\beta_2/2$ and  $\gamma^{'}=\beta_2^{'}/2.$
\end{proof}

%

\subsection{Additional Experiment Figures
}

\end{document}